\documentclass[final,5p,times,twocolumn]{elsarticle}
\usepackage{algorithm}
\usepackage{algorithmic}
\usepackage{amsmath}
\usepackage{amssymb}
\usepackage{array}
\usepackage{bm}
\usepackage{booktabs}
\usepackage{cases}
\usepackage{graphicx}
\usepackage{multirow}
\usepackage{url}
\usepackage{xcolor}
\usepackage{makecell}
\usepackage{pifont}
\usepackage{subfig}
\usepackage[colorlinks,
            linkcolor=red,
            anchorcolor=black,
            citecolor=black]{hyperref}

\newcommand{\eg}{\textit{e.g.},~}
\newcommand{\ie}{\textit{i.e.},~}

\journal{Information Fusion}
\begin{document}
\begin{frontmatter}

\title{Exploring Fusion Strategies for Accurate RGBT Visual Object Tracking}

\author[a]{Zhangyong Tang}
\author[a]{Tianyang Xu}
\author[a]{Hui Li}
\author[a]{Xiao-Jun Wu\corref{correspondingauthor}}
\author[a]{XueFeng Zhu}
\author[b]{Josef Kittler}
\address[a]{Jiangsu Provincial Engineering Laboratory of Pattern Recognition and Computational Intelligence, \\ School of Artificial Intelligence and Computer Science, Jiangnan University, \\ 214122, Wuxi, China \fnref{address1}}
\address[b]{The Center for Vision, Speech and Signal Processing, University of Surrey, Guildford, GU2 7XH, UK. \fnref{address2}}

\cortext[correspondingauthor]{Corresponding author email: wu\_xiaojun@jiangnan.edu.cn}

% \author{Zhangyong~Tang,
%         Tianyang~Xu,
%         Hui~Li,
%         Xiao-Jun~Wu*,
%         Xue-Feng~Zhu,
%         and~Josef~Kittler~\IEEEmembership{Life~Member,~IEEE}% <-this % stops a space
% \thanks{Z. Tang, T. Xu, H. Li, X. Zhu, and X.-J. Wu (Corresponding Author) are with the School of Artificial Intelligence and Computer Science, Jiangnan University, Wuxi, P.R. China. (e-mail: \{zhangyong\_tang\_jnu, tianyang\_xu, hui\_li\_jnu, xuefeng\_zhu95, xiaojun\_wu\_jnu\}@163.com)}
% \thanks{J. Kittler is with the Centre for Vision, Speech and Signal Processing, University of Surrey, Guildford, GU2 7XH, UK. (e-mail: j.kittler@surrey.ac.uk)}
% \maketitle
%
% \sloppy
\begin{abstract}
We address the problem of multi-modal object tracking in video and explore various options of fusing the complementary information conveyed by the visible (RGB) and thermal infrared (TIR) modalities including pixel-level, feature-level and decision-level fusion. 
Specifically, different from the existing methods, paradigm of image fusion task is heeded for fusion at pixel level.
Feature-level fusion is fulfilled by attention mechanism with channels excited optionally.
Besides, at decision level, a novel fusion strategy is put forward since an effortless averaging configuration has shown the superiority.
The effectiveness of the proposed decision-level fusion strategy owes to a number of innovative contributions, including a dynamic weighting of the RGB and TIR contributions and a linear template update operation.
A variant of which produced the winning tracker at the Visual Object Tracking Challenge 2020 (VOT-RGBT2020).
The concurrent exploration of innovative pixel- and feature-level fusion strategies highlights the advantages of the proposed decision-level fusion method.
Extensive experimental results on three challenging datasets, \textit{i.e.}, GTOT, VOT-RGBT2019, and VOT-RGBT2020, demonstrate the effectiveness and robustness of the proposed method, compared to the state-of-the-art approaches. 
Code will be shared at \textcolor{blue}{\emph{https://github.com/Zhangyong-Tang/DFAT}.
}
\end{abstract}

\end{frontmatter}
%
% \IEEEpeerreviewmaketitle
%

\section{Introduction}
Visual object tracking, aiming at predicting the state of a target of interest in video sequences, is a fundamental task in computer vision. 
It has received considerable attention over the past decades owing to its wide range of applications, such as robotics \cite{IF-robotics}, video surveillance \cite{IF-surveillance} and pedestrian tracking \cite{IF-pedestrian}, to name a few. 
Generally, visible spectrum (RGB) and thermal infrared (TIR) tracking are two sub-tasks in visual object tracking. 
RGB data contains more detailed perceptual information about the scenario, but the tracking performance is sensitively affected by challenging appearance variations, such as illumination, blur, and occlusion. 
On the contrary, the TIR channel is more robust against lighting conditions, but with less discriminative power owing to thermal crossover and insufficient texture details. 
Hence, there is a major scope for taking advantage of the complementary information provided by these two modalities to design a robust tracker. 
This multi-modal approach is called RGBT tracking.
\begin{figure}
	\begin{center}
		\includegraphics[width=1\linewidth]{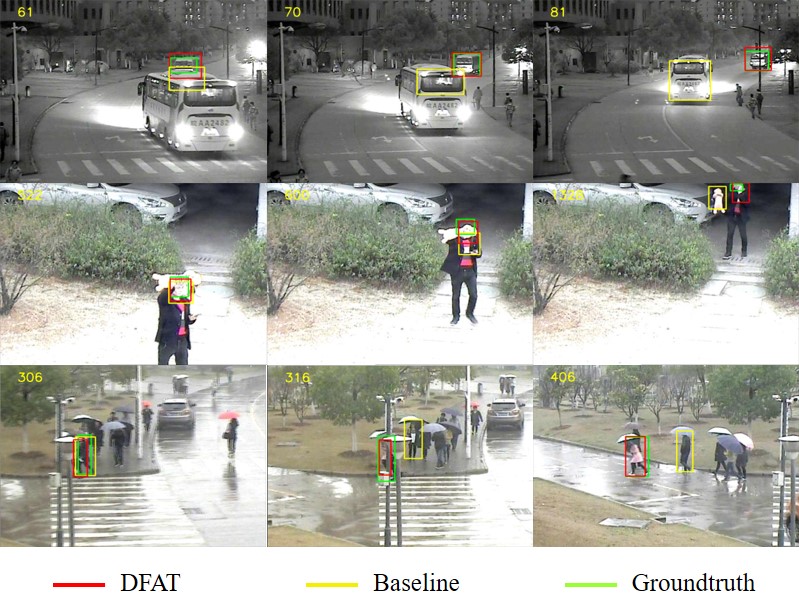} 
	\end{center}
	\caption{Qualitative comparison between our method (DFAT) and Baseline, with ground truth on the VOT-RGBT2019 \cite{VOT2019} dataset. Respectively, frames sampled from video sequences \emph{bus6}, \emph{face1}, and \emph{woman89} are shown in the first, second and third rows of the figure. Here 'Baseline' means SiamRPN++ (RGBT), which is the multi-modal extension of SiamRPN++ \cite{SiamRPN++} in RGBT tracking.}
	\label{fig:results_image_figure}
\end{figure}
\par With the development of RGB and thermal infrared all-in-one devices, RGBT tracking has become an important research topic in the visual tracking community. 

In general, the existing approaches mainly conduct the cross-modal fusion of the RGB and TIR content at pixel \cite{mfDimp} or feature level \cite{MANet, DAPNet}. 
Pixel-level fusion directly concatenates the information from one modality with the other at the tracking input stage.
This coarse fusion strategy endows the same importance for the salient and low-frequency clues existed in both RGB and TIR data.
However, following the tracking framework, salient parts are supposed to be more important than low-frequency parts.

%This strategy is very effective with hand-crafted features.
%However, considering the misalignment caused by manually calibration, fusion at this stage will generate polluted source data for the following steps and  especially not suitable for an end-to-end paradigm.
%However, ignoring the misaligned multimodal images, it is not suitable for deep neural networks, especially in an end-to-end paradigm.

In contrast, feature-level fusion performs the fusion processing in an embedded way, enabling integrated feature extraction by deep neural networks.
Typically, the neural networks are trained on one dataset and tested on other datasets, \eg trained on RGBT234 \cite{RGBT234} and tested on GTOT \cite{GTOT}. 
However, in the VOT-RGBT challenge \cite{VOT2019}, GTOT is the largest RGBT dataset permitted for model training.
The number of video sequences in the other two RGBT datasets, \ie LITIV \cite{LITIV} and OSU-CT \cite{OSU-CT}, is less than 20 and no more than 20000 image pairs are contained in these three datasets.
% There are approximately 40000 frames in these datasets. 
Consequently, since there is an inadequate amount of training data, it is impossible to guarantee the network would have a sufficient capacity to handle general scene content after training on the available RGBT sequences. 
The deep learning trackers designed by using a small number of training videos (limited to 65 sequences in total) will have to be lightweight, with a high risk of over-fitting. 
The resulting solutions are unable to deliver discriminative representations for both RGB and TIR images. 
In conclusion, the lack of annotated RGBT tracking data limits the representation capacity of the networks and compromises the efficiency of the embedded-level feature fusion. 

Different from the above strategies, fusion at decision level has not been extensively studied in RGBT tracking. 
Traditional correlation filter (CF) \cite{KCF} based trackers, like \cite{kld}, adopt only hand-crafted features to calculate the responses for both modalities, which are then directly fused to predict the target location. 
%In this paradigm, the tracker in \cite{kld}, which is based on the correlation filter (CF) \cite{KCF}, adopts only hand-crafted features to calculate the responses for both modalities, which are then fused to predict the target location.
This approach is less effective when the target undergoes severe appearance variation.
In addition, advanced RGB trackers, such as \cite{LADCF} and \cite{GFSDCF}, are equipped with pre-trained feature extractors to get better embeddings and improve the tracking performance. 
Although it has recently been demonstrated that single modality trackers have excellent properties after allocated superior feature extractors, extending their success to RGBT tracking through decision-level fusion is still a challenge.
Since there exists a gap between the imaging mechanism of RGB and TIR modalities.

Considering the aforementioned concerns that fusion at all these three stages have their shortcomings, we investigate three fusion strategies (fusion at pixel-level, feature-level, and decision-level) and perform a comprehensive comparison of their different characteristics in the RGBT tracking scenario. 
Specifically, at pixel-level, inconsistent with the naive addition/concatenation operator used in existing methods, we introduce a general image fusion method to get the fused image with different focuses lying on the salient and low-frequency parts.
For online feature-level fusion, adaptive fusion weights are employed to merge the information from the two spectra. 
Before that, to eliminate the feature noise and redundancy, different from the Weighted Random Selection strategy (WRS) \cite{WRS} used in \cite{DAPNet}, an adaptive selection method is conducted to suppress the channels with lower discrimination by channel attention mechanism. 
For fusion at decision level, originally, the results from both classification and regression branches are simply averaged for the comparison with fusion at the other two stages.
However, under the constraint of limited amount of data, it is unexpected that although conducted by a primitive fusion strategy, decision-level fusion has already shown its superiority against fusion at pixel and feature levels.
Furthermore, decision-level fusion has already been verified effective in other multi-modal tasks \cite{IF-decision-level}.

Drawing on this, we propose an adaptive decision-level fusion strategy for accurate RGBT tracking (DFAT) in an end-to-end framework, whose tracking results are intuitively presented in Fig. \ref{fig:results_image_figure}.
To address the problem of insufficient quantity of annotated RGBT data for model training, we use only the RGB sequences to train the tracking network offline and perform cross-modal fusion in the online tracking stage. 
More specifically, we use publicly available large datasets of annotated RGB images, \eg COCO \cite{COCO} and YouTube-VOS \cite{YouTube-VOS} for network training. 
For decision-level fusion, based on the observation that the response scores of RGB images are higher than those of TIR images in most sequences, which is caused by the optical difference between RGB and TIR imaging principle, we design a novel strategy that focuses on suppressing the bias between these two modalities. 
Fig. \ref{bias-score} also gives an intuitive illustration of the bias.
To this end, we assign an adaptive weight, which is specified for debiasing, to each modality to harmonise the complementary multi-modal appearance.
% and the tracking performance is significantly improved.
% By assigning an adaptive weight to each modality, the fusion performance with the resulting normalization of the scores of the two modalities is significantly improved. 
% To retain the consistency of data, we shrink the scores to a magnitude similar to the original scores.
As demonstrated by the experiments, the proposed DFAT effectively exploits the synergies of the two modalities, delivering superior tracking performance, as compared to other fusion strategies.
% and the tracking results shows in Fig. \ref{fig:results_image_figure}.

\par Our contributions can be summarized as follows:
\begin{itemize}
	\item As potential alternatives, we concurrently investigate three innovative pixel-level, feature-level and decision-level fusion methods for RGBT tracking, with the results showing that the decision-level fusion is significantly superior. 
	\item In the context of the limited availability of annotated RGBT datasets for training a multi-modal tracker, using the Siamese network trained on RGB videos for embedding both modalities, RGB and TIR, we propose an online adaptive decision-level fusion method to excavate the hidden complementaries.
	A variant of the proposed tracker won the VOT-RGBT (2020) Challenge.
	\item Based on the experimental phenomenon that the gap between RGB and TIR modalities leads to classification bias, which is displayed in Fig. \ref{bias-score}, a dynamic weighting methodology of the RGB and TIR contributions is developed at decision level.  
	\item The experimental results on the VOT-RGBT2019 and GTOT datasets demonstrate that our method, DFAT, achieves promising performance. Moreover, on the VOT-RGBT2020 dataset \cite{VOT2020}, DFAT defines the new state-of-the-art in RGBT tracking.
\end{itemize}

%The key innovations of the proposed decision-level fusion approach are two mechanisms: i) a dynamic weighting of the RGB and TIR contributions to the classification branch, and ii) a classification branch scaling to balance the relative contributions of the classification and regression branch response maps.
\par The rest of this paper is organised as follows. 
The related work and the proposed method are discussed in Section II and Section III.
Extensive experiments and comprehensive analysis of the experimental results are reported in Section IV. 
Finally, a conclusion on the research findings is drawn in Section V.

\section{Related work}
As mentioned above, RGBT tracking, aiming at fusing the complementary information between RGB and TIR modalities, is a sub-task of visual object tracking. 
In this section, we briefly introduce the relevant tracking approaches with single and multiple modalities.

\subsection{Tracking with Single Modality}
RGB tracking is the most fundamental sub-task in visual object tracking \cite{LADCF,GFSDCF,LSDCF,ZhuTCSVT,SiamMask}.
Among its numerous modelling techniques, trackers based on Siamese networks are the widely studied in the recent deep learning paradigm. 
SiamFC \cite{SiamFC}, which is the seminal work that introduced Siamese networks into object tracking, uses an end-to-end network trained offline to learn a general similarity metric without a fully connected layer. 
To obtain more accurate predictions for the target bounding box, Region Proposal Network (RPN) \cite{RPN} is firstly utilized in SiamRPN \cite{SiamRPN}. 
The above-mentioned trackers only use features from the output of one specific CNN layer, while features from different CNN layers exhibit unequal spatial resolution and semantic extent. 
To mitigate this problem, C-RPN \cite{CascadedRPN} uses cascaded RPN blocks to combine high-level semantics with low-level spatial information. 
Besides, to address the issue that Siamese-based methods can not make full use of deeper and wider architectures, SiamRPN++ \cite{SiamRPN++} presents a spatially-aware sampling strategy designed to exhibit adaptation between the deeper network and the basic Siamese formulation. 
To this end, ResNet-50 \cite{Resnet50} is employed as the backbone of SiamRPN++, achieving improved performance compared with AlexNet backbones.

\par TIR tracking is also a hot topic in visual object tracking \cite{HSSNet, MMNet}.
Different from Siamese-based RGB trackers, most TIR trackers model the target with only handcrafted features rather than deep networks. 
Many TIR trackers, as in \cite{SRDCFir}, are simple extensions of RGB trackers.
For example, SRDCFir extends SRDCF \cite{SRDCF} to TIR tracking by introducing motion features while using the same modelling technique. 
EBT \cite{EBT} uses a trained Support Vector Machine (SVM) classifier to re-rank the edge boxes, thus obtaining instance-specific proposals. 
Similarly, DSLT \cite{DSLT} combines motion features with edge and Histogram of Oriented Gradients (HOG) \cite{HOG} features, achieving real-time tracking in the Fourier domain. 
In principle, TIR trackers seldomly use neural networks to extract features, since there is insufficient data to train a network to learn a satisfactory representation for the TIR modality. 
To this end, ECO-tir \cite{ECO-tir}, based on the RGB tracker ECO \cite{ECO}, generates a synthetic TIR dataset from existing RGB datasets by a Generative Adversarial Network (GAN) \cite{GAN}. 
The modality-specific information learned from the generated TIR data contributes to its (\cite{mfDimp}) favourable results on VOT-RGBT2019 \cite{VOT2019} dataset. 
To enhance the robustness, MLSSNet \cite{MLSSNet} regards tracking as a matching problem and proposes a multi-level similarity model, which incorporates global and local similarities. 
Besides proposing a tracking algorithm, MLSSNet also provided the first large scale TIR dataset, containing 430 video sequences. 
Similarly, MMNet \cite{MMNet} uses a multi-task architecture to integrate the class-level and fine-grained features. 
The authors also released a dataset consisting of 30 classes and over 1100 video sequences for TIR tracking. 

\begin{figure*}
	\begin{center}
		\includegraphics[width=1\linewidth,height=0.4\linewidth]{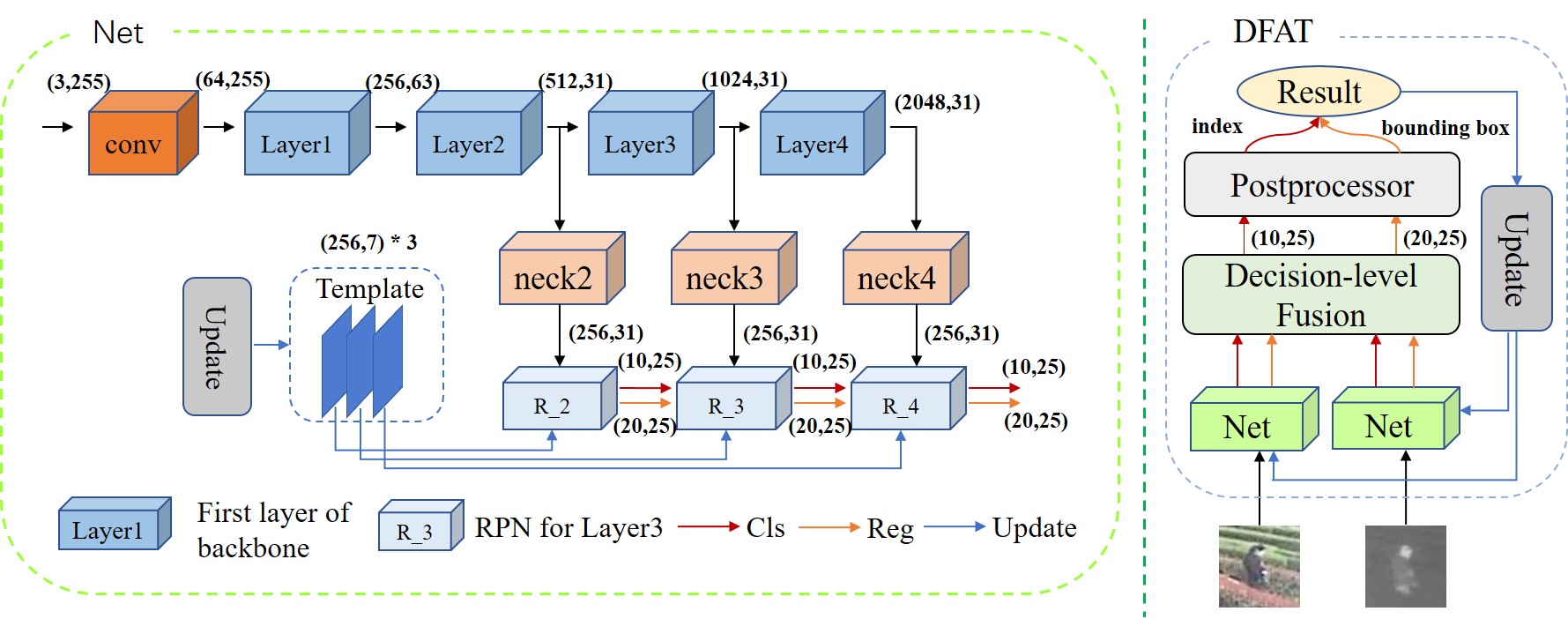} 
	\end{center}
	\caption{Illustration of the proposed DFAT method. Here 'Cls' and 'Reg' represent classification and regression branches respectively. The left part describes the architecture of the feature extraction network ('Net'). The outputs from three convolutional layers are first projected into a common space by neck modules and then input into the corresponding RPN blocks. The features from different RPN blocks are modulated by adaptive weights within 'Net'. An overview of the proposed method (DFAT) is on the right. The outputs of 'Net' modules from both modalities are fused at the 'Decision-level Fusion' module by Eq.~(\ref{fuse}). And the final result is obtained after the 'Postprocessor'. We update the template every 10 frames. For convenience, the dimensionality of feature representation is defined by the number of channels and the spatial resolution. Thus, the shorthand notation (3,255) signifies that the size of the input is 3x255x255.
	%The final outputs of the two modalities are fused at the decision-level to obtain the results for the fused 'CLS' and the fused 'REG' (Eq.~(\ref{fuse})). The final 'Result' is obtained from 'REG' according to the most similar anchor chosen by 'CLS'. We update the template every 10 frames. For convenience, the dimensionality of features is defined by the number of channels and the spatial resolution. Thus, the shorthand notation (3,255) signifies that the size of the input is 3x255x255.
	%with different fusion mediums implanted for different branches, from which the result is obtained and employed in the template 
	}
	\label{pipeline}
\end{figure*}

\subsection{Tracking with Multiple Modalities}
Nowadays, tracking with multi-modal inputs draws increasing attention, such as RGBT \cite{IF-rgbt-review} and RGB-Depth tracking \cite{IF-RGBD}.
And the former, RGBT tracking, is thoroughly discussed in this paper.
Joint visual object tracking in RGB and TIR modalities has been proposed to create more robust solutions in practical scenarios .
The core issue of multi-modal tracking is to merge the information from both modalities efficiently. 
In this subsection, RGBT trackers will be introduced according to the fusion strategies.

In general, fusion at pixel level is fulfilled in an add-up fashion. 
\cite{2006AVE} uses averaging technique and achieves satisfactory visualisation while \cite{pixel-fusion} employs Dual-Tree Complex Wavelet Transforms (DT-CWT) \cite{DT-CWT}, obtaining higher performance in its quantitative experiments.
%no short name for 2006AVE and pixel-fusion
In recent years, seldom trackers explore the fusion stage at pixel-level because of the existence of misalignment between RGB and TIR image pixels.
A simple concatenation operator is utilized for pixel-level fusion in \cite{mfDimp} to manifest the superiority of its fusion strategy.

The most widely used fusion strategy is applied at the feature level. 
%For trackers without training the end-to-end deep architecture, features extracted by neural networks or hand-crafted functions are also fused in an add-up fashion. 
%However, these two modalities contribute diversely in practice. 
%With the development of fusion techniques, the method with adaptive weights is proposed in \cite{GTOT} by calculating reconstruction residues in the Bayesian filtering framework. 
Adaptive weights are obtained in \cite{GTOT} by calculating reconstruction residues in the Bayesian filtering framework.
Similarly, the weights in \cite{lan2018robust} are acquired by the learned classifiers for each modality under the max-margin principle. 
However, intuitively, feature-level fusion should be accomplished deeper in an end-to-end framework. 
To this end, \cite{Chenglong2018Fusing} proposes a lightweight fusion sub-network, which consists of two convolutional layers, to accomplish feature selection for accurate RGBT tracking.
FANet \cite{FANet}, similarly, utilizes multi-layer features, whose spatial resolutions are unified by max pooling. 
To improve the representation capacity, an adaptive fusion sub-network is applied to suppress channels with less reliability. 
DAPNet \cite{DAPNet} simultaneously integrates all the chosen layers and modalities by using an adaptive fusion module recursively, achieving feature pruning by global average pooling to reduce redundancy and noise.
Further, DAFNet \cite{DAFNet} integrates the adaptive fusion sub-network from \cite{FANet} and realizes quality-aware fusion in each layer between the two modalities. 
To achieve fine-grained fusion, MANet \cite{MANet} divides features into three categories (modality-specific, modality-shared, and instance-specific), exploiting a more exhaustive feature-level fusion approach to get robust feature representation by multiple adaptors.
%The above end-to-end trackers are all on the basis of MDNet\cite{MDNet}.
On account of the efficiency of Siamese structure in RGB tracking, SiamFT \cite{SiamFT} primarily brings in this structure to RGBT tracking and learns fusion weights from the medium-term calculated response maps.
Furthermore, based on dynamic Siamese network \cite{DSiam}, a multi-layer fusion strategy is adopted in DSiamMFT \cite{DSiamMFT} to enrich the semantic information of target in both modalities. 
To integrate the temporal clues, CMPP \cite{CMPP} achieves pattern propagation spatially for inter-modal and temporally for intra-modal by cross-modal attention.
A multi-branch architecture is employed in CAT \cite{CAT} for target appearance modelling in allusion to the modality-specific and modality-shared challenges, which totally count to 5. 
Unlike DAPNet \cite{DAPNet} and DAFNet \cite{DAFNet} keep only the fused features retained, TFNet \cite{TFNet} designs a trident architecture for better excavating the modality-specific information.
Constraint to the limited RGBT data, the trackers aforementioned are all equipped with lightweight feature extractors. 
To make better use of deep neural networks, one option is to generate TIR data from existing RGB data \cite{ECO-tir}.  
%It has been shown in \cite{mfDimp} that using the dataset synthesized from GOT-10k \cite{GOT-10k} improve the tracking performance.
It has been shown in \cite{mfDimp} that, with the dataset synthesized from GOT-10k, better performance is obtained.
However, we argue that synthesized data can not adequately represent the real TIR data in terms of content and texture. 
Besides, there are some low-level topics which regard the RGBT tracking as a downstream task to verify their robustness and generalization like RFN-Nest \cite{IF-RFN-Nest}
%Compared with fusion at feature-level, decision-level fusion has not been extensively studied.

Lastly, seldom trackers pay attention to the fusion at the decision level.
\cite{IPT2019} and \cite{kld} use KL Divergence to estimate the reliability and achieve adaptive fusion after the response maps of both modalities are available.
\cite{mfDimp} sums up the scores from different modalities roughly and only obtains similar results compared with pixel-level fusion.
Once the response maps are calculated from appearance trackers, \cite{JMMAC} learns local and global weights from the original image patch via a lightweight network.
Then the decision-level fusion is applied in a pixel-wise manner.

However, in this work, we explore fusion at all the three locations.
For pixel-level fusion, unlike \cite{mfDimp} and \cite{JMMAC}, we adopt an advanced image fusion method, MDLatLRR\cite{MDLatLRR}, to separate the detailed and salient information from paired RGBT images. 
The salient information is averaged directly while the detailed information is fused by nuclear-norm, and both of them are integrated to get the final fused image. 
For fusion at feature level, a feature selection method is designed based on channel attention mechanism while this in \cite{DAPNet} and \cite{TFNet} is carried out by WRS. 
Furthermore, spatial attention mechanism is also considered for the cross-modal fusion.
Decision-level fusion, experimentally, is proved to be the best under the background of our method.
Specifically, our adaptive fusion weights are learned based on the hypothesis that there is supposed to be a gap between the training RGB and testing TIR data, which is illustrated in Fig. \ref{bias-score}.
Compared with JMMAC, VGGNet is used as the feature extractor in JMMAC while ResNet-50 in our method, which is demonstrated to be better and can extract features with more discrimination most of the time.
Besides, the weight matrix is learned from input image patches in JMMAC and our weight scalar is acquired conveniently and rapidly from the classification results.
A more comprehensive comparison of multi-modal fusion is contained in this work since fusion at the pixel level is not involved in JMMAC.
%Differently, we aim to enhance the multimodal tracking performance by focusing on fusion at the decision level, given the limited size of the RGBT training set.
%We propose an adaptive fusion strategy to better utilize the complementary characteristic conveyed by the two modalities. 
%The efficiency of the decision-level fusion has already been demonstrated in \cite{kld} and \cite{JMMAC}. \cite{kld} uses KL Divergence to fuse the response maps. 
%However, \cite{kld} is based on CF where the desired label distribution is Gaussian, which is particularly suited for use with the KL Divergence measure. It is less appropriate for response maps generated by siamese networks. 
%Besides, ECO \cite{ECO}, which is also a CF-based tracker, is employed as the baseline of \cite{JMMAC}.

\section{Proposed Method}
In this section, we first briefly introduce our baseline method SiamRPN++ \cite{SiamRPN++}, which is a single modality tracker with an end-to-end training framework. 
Then, we will discuss our decision-level fusion strategy as well as fusion at pixel and feature levels for a comprehensive comparison.

\subsection{Baseline Tracker}
The framework of SiamRPN++ architecture is displayed in the left of Fig. \ref{pipeline}.
% Attentively, the 'Update' block isn't contained in SiamRPN++.
In the tracking stage, a typical Siamese tracker fixes the template features from the first frame and keeps it unchanged for the rest of tracking frames.
When a new frame (instance) comes, SiamRPN++ first extracts multi-layer feature maps (layer modules) and performs channel reduction (neck modules). 
Once the feature embeddings are extracted, the results of classification (Cls) and regression (Reg) branches from multiple RPN blocks (R\_2, R\_3, R\_4) can be calculated.
In the baseline tracker, the multi-modal results are directly averaged for fusion.
After that, the general 'Postprocessor' is followed to acquire the tracking result from the fused classification and regression results. 
Specifically, for the classification maps, a softmax operator is employed to convert the original foreground/background response scores into probabilities with values from the interval $[0,1]$, intuitively measuring similarities. 
%For trackers without 'Decision-level Fusion' module, the multi-modalities 
%For the classification branch, a softmax operator is employed to convert the original foreground/background response scores into probabilities with values from the interval $[0,1]$, which makes measuring similarity more intuitive (CLS). 
As for the regression branch, the result returned is transferred to the bounding box since the offsets between anchors and ground truth are learned from the given labels.
Consequently, the index of the highest value among the normalized classification scores is selected and its corresponding bounding box in the regression branch is considered as the final tracking result.

% The main contributions of SiamRPN++ involve two parts, \ie online tracking and offline training. 
As for online tracking, SiamRPN++ uses multi-layer features to improve the reliability of tracking.
The computation process is defined in the following formulation:
\begin{equation}
\left\{
\begin{aligned}
\textbf{\emph{C}}=&\sum\limits_{i} \alpha_i\textrm{R}^\textrm{C}_i(f_{i}(\textbf{\emph{X}}), f_{i}(\textbf{\emph{Z}})),\\
\textbf{\emph{B}}=&\sum\limits_{i} \beta_i\textrm{R}^\textrm{B}_i(f_{i}(\textbf{\emph{X}}), f_{i}(\textbf{\emph{Z}}))
\end{aligned}
\right. .
\end{equation}
where $\textbf{\emph{X}}$ and $\textbf{\emph{Z}}$ are the image patches from an instance and the exemplar. 
$f_{i}$ represents the output of the $i$th neck module after the $i$th backbone layer. 
$\textrm{R}_i^\textrm{C}$ and $\textrm{R}_i^\textrm{B}$ denote the classification and bounding box regression branches respectively in the $i$th RPN block. 
% $\textbf{\emph{C}}$ and $\textbf{\emph{B}}$ are the output of the classification and regression branches, respectively. 
% $w_i$, which includes two parameters 
$\alpha_i$ and $\beta_i$ are the weights associated with the corresponding branches for the $i$th layer, $\textrm{R}_i^\textrm{C}$ and $\textrm{R}_i^\textrm{B}$, which are also learned offline. 
% $\alpha_i$ is learned for $\textbf{\emph{C}}_i$ while $\beta_i$ is for $\textbf{\emph{B}}_i$.
% $\textbf{\emph{I1}}$ and $\textbf{\emph{I2}}$ are the two parameters specifying  the input to the RPN blocks. 
As shown in Fig. \ref{pipeline}, SiamRPN++ uses the features from layer 2-4 and fuses them by utilizing multiple RPN blocks. 
In addition, depth-wise cross-correlation is adopted to solve the parameter distribution imbalance caused by the increased dimensions.

\par As for offline training, SiamRPN++ employs a sampling strategy with translations to mitigate the constraint that Siamese trackers are unable to take advantage of deeper networks like ResNet-50 \cite{Resnet50}. 
Specifically, the basic formulation of the Siamese framework imposes a strict translation invariance, and unfortunately, the padding operation is incompatible with this constraint. 
Moreover, padding is an essential mechanism to match the input size specification for deep networks. 
Based on an in-depth analysis, SiamRPN++ devises a simple sampling method that preserves the same spatial resolution for layer 2, layer 3, and layer4, and works effectively. 
As a result, deep networks, such as ResNet-50 \cite{Resnet50} and VGGNet \cite{VGGNet}, achieve improved performance compared to the original AlexNet \cite{AlexNet}, as shown in \cite{CIR}.

\subsection{Fusion Mechanism}

\begin{figure}
	\begin{center}
		\includegraphics[width=1\linewidth]{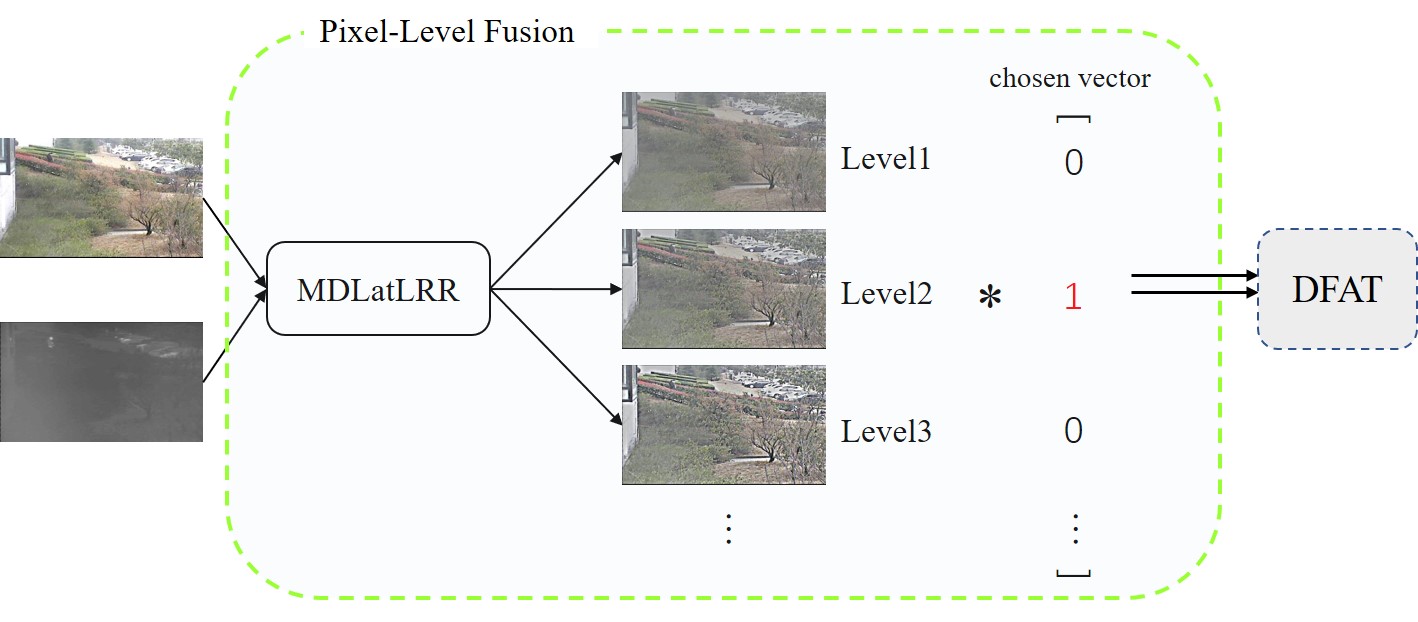} 
	\end{center}
	\caption{Introduction of pixel-level fusion achieved by MDLatLRR. The paired RGB and TIR images are first fed into the MDLatLRR module with the chosen vector manually determinated and the fused output is then sent to the tracking network.}
	\label{pixel-level}
\end{figure}

We deliver a brief introduction to our techniques of fusion at pixel, feature and decision levels respectively.
A novel decision-level fusion strategy, since fusion at decision level performs considerable better results than fusion at pixel and feature levels, is then described in this subsection and our pipeline is given in Fig. \ref{pipeline}. 

\par \textbf{Pixel-level fusion:} For pixel-level fusion, we do not concatenate the image matrices directly as \cite{mfDimp} do. 
Instead, we choose MDLatLRR \cite{MDLatLRR}, which has been proved to be efficient, to fuse the RGB and TIR images. 
MDLatLRR uses multi-level LatLRR-based \cite{LatLRR} decomposition to detach the detailed and salient information from both RGB and TIR images. 
The low-frequency parts are fused by weighted averaging while adaptive weights are learned for the fusion of detail parts based on nuclear-norm. 
Fig. \ref{pixel-level} shows the minutiae of pixel-level fusion. 
The chosen vector is set in advance to acquire the fused image at a particular level and only one element is supposed to be activated. 
Particularly, the result from level2 is selected as input in Fig. \ref{pixel-level}. 

\par
\begin{figure}
	\begin{center}
		\includegraphics[width=1\linewidth]{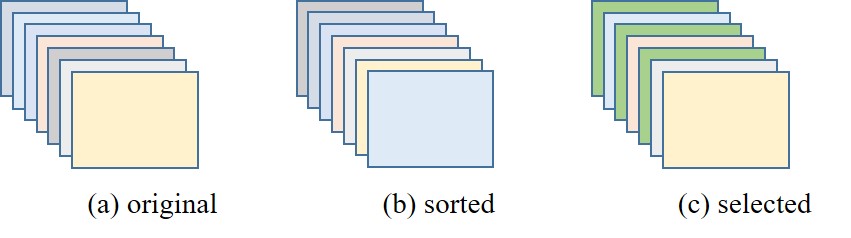} 
	\end{center}
	\caption{Illustration of the process of feature selection for the fusion of the two modalities at the feature level, which is embedded in after the neck module. Here the channels highlighted with dark green represent those with less discrimination and are further suppressed.
% 	We change the value of a particular channel (highlighted in dark green) to zero and keep the order unchanged.
}
	\label{feature-level}
\end{figure}

 \textbf{Feature-level fusion:} Fusion at feature level is carried out using attention mechanism. 
Generally, feature selection is executed through channel mechanism within each modality.
After that, spatial and channel attention, combined with normalization, is further employed for cross-modal fusion.
% In particular, either spatial or channel attention is exploited to obtain the attention matrix by average or max pooling. 
% The attention learned from RGB and TIR features is used for fusion after normalization. 
In particular, considering the noise and the level of redundancy in deep features, which are discussed in detail in \cite{GFSDCF}, we apply feature selection, which involves two parts, \ie channel saliency measurement and channel selection, to achieve adaptive fusion. 
For assessing the channels, we first employ adaptive pooling to indicate the significance of the individual channels.
In other words, the discriminatory information conveyed by each channel is represented by an attention value. 
For channel selection, as shown in Fig. \ref{feature-level}, we first sort the channels by the learned representative scalar in the descending order of its values. 
We retain the top 80\% of channels based on the significance score and set the others to zero.
%without intermingling their gradation. 
% The subsequent steps are the same as before (without feature selection). 
Subsequently, spatial or channel attention is conducted by adaptive pooling and results in representative vectors, from which the final fusion scalars are computed by mean operator.
The operations outlined above are applied after the neck module, as the projection to a low-dimensional space can decrease noise and suppress redundancy. 
\par \textbf{Decision-level fusion:} To address the issue raised by the lack of RGBT data, especially for the VOT-RGBT challenge, we investigate the merits of fusing at the decision level.
Ordinarily, RGB and TIR images are simultaneously fed into the tracking network, with two pairs of outputs received from the RPN blocks.
The averaged results of classification and regression branches are then fed into the 'Postprocessor'.
Specifically, the scores of the classification branch are normalized to probabilities by a softmax operator. 
Finally, the bounding box is determined based on the maximum value of the normalized scores. 
The experiments suggest that fusion before normalization works better as shown in TABLE \ref{results_decision_table}. 
For this reason, in this paper, we focus on fusion before normalization. 
Although fusion at this stage yields considerable improvements, we find that the scores produced by RGB images are always greater than the TIR image scores as shown in Fig. \ref{bias-score}\textcolor{red}{(a)} and \ref{bias-score}\textcolor{red}{(c)}, which means that weighting the scores directly may produce inherently biased results. 
% In Fig. \textcolor{green}{\ref{databias-score}}, one point means the maximum of scores of a certain frame.

% In our approach, images from both modalities are identically input to the network. 
% To design an adaptive fusion mode for the results from the classification branch, we analyze the bias in the output scores intensively. 

\begin{figure}
	\begin{center}
		\includegraphics[width=1\linewidth]{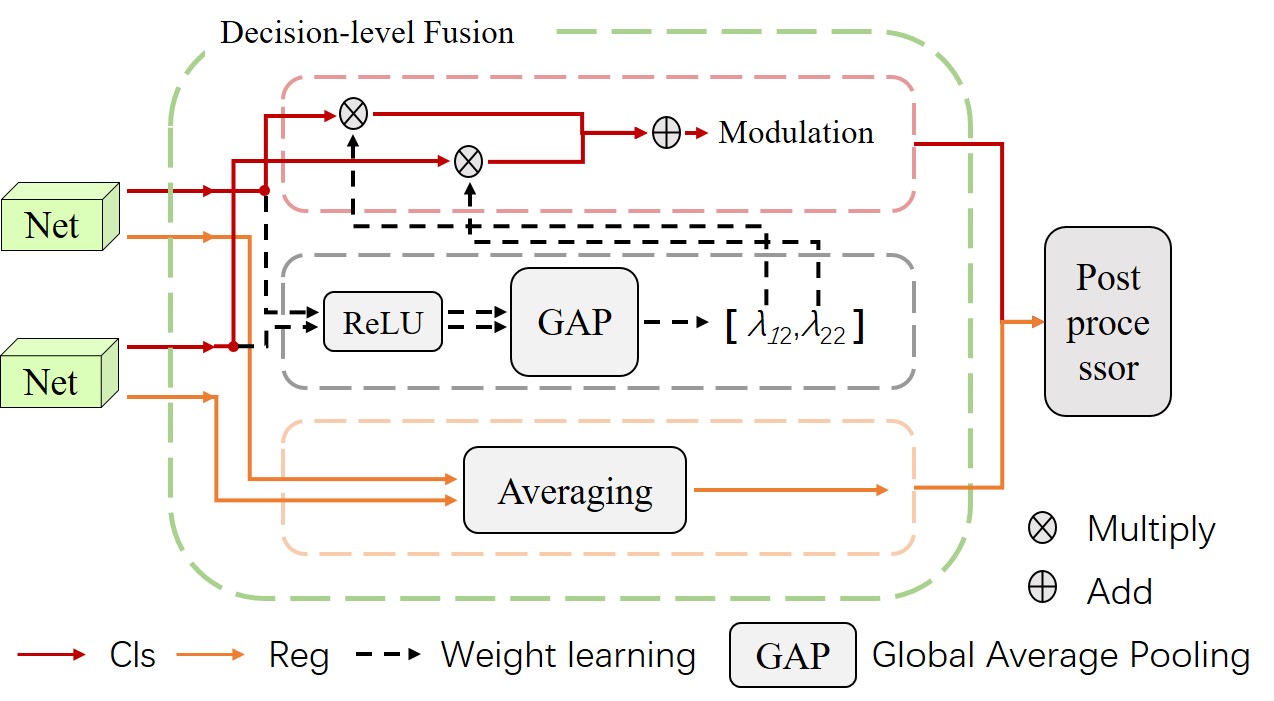} 
	\end{center}
	\caption{Details of our decision-level fusion module. 'Net' and 'Postprocessor' have already been introduced in Fig. \ref{pipeline} and Section III as parts of the pipeline. The results from the regression branches are averaged while the results from the classification branches are adaptively weighted. Here 'Modulation' represents two numerical operations, \ie shrink and scale, which are detailed in the description of our decision-level fusion method. Eq.~(\ref{fuse}) is the mathematical description of our decision-level fusion.}
	\label{decision-level}
\end{figure}

Fig. \ref{decision-level} shows the details of our decision-level fusion method.
Since we select the bounding box according to the maximum of the normalized response scores (classification branch), we first use ReLU (Rectified Linear Unit) activation \cite{ReLU} to suppress the locations with a response below zero, which is less likely to be the selected index paired with the highest similarity.
Besides, we amplify the scores of RGB and TIR images to the same magnitude by multiplying them by the average value, which is calculated by GAP (Global Average Pooling) \cite{GAP}, to alleviate the bias.
% After weighting, the adding operator is introduced. 
The final result produced by the classification branch is weighted by a factor consisting of two parts, \ie shrink and scale.
Specifically, to maintain the data consistency, we shrink the zoomed-in data to a magnitude similar to the original scores by dividing by the sum of average values. 
To further adjust the scores after shrinkage, we introduce a hyper-parameter called the scaling factor, which balances out the relative contributions of the classification and regression branches.
%is supposed to increase the distance between the normalized scores and contribute to accurate index selection.
More details of the above operators are presented in the experimental section. 
As for the regression branch, we obtain the fused result in a similar way. 
% However, as explained in \cite{RGBT234}, there is a discrepancy between the bounding box of the RGB image and its corresponding TIR image. 
% In this paper, we address this issue by manually assigning the weights for the regression branches of the two modalities. 

The decision-level fusion process is formulated as follows:
\begin{equation}\label{fuse}
\begin{split}
\textbf{\emph{B}}_F & = (\lambda_{11} * \textbf{\emph{B}}_\textrm{RGB} + \lambda_{21} * \textbf{\emph{B}}_\textrm{TIR})/(\lambda_{11} + \lambda_{21})\\
\textbf{\emph{C}}_F & = s * (\lambda_{12} * \textbf{\emph{C}}_\textrm{RGB} + \lambda_{22} * \textbf{\emph{C}}_\textrm{TIR})/(\lambda_{12}+ \lambda_{22})\\
\lambda_{12} & = M_{\textbf{\emph{C}}_\textrm{TIR}>0}(\textbf{\emph{C}}_\textrm{TIR})\\
\lambda_{22} & = M_{\textbf{\emph{C}}_\textrm{RGB}>0}(\textbf{\emph{C}}_\textrm{RGB})
\end{split}
\end{equation}
%\lambda_{12} & = \frac{\|C_R\|^2+\|C_I\|^2}{\|C_R\|^2}\\
%\lambda_{22} & = \frac{\|C_R\|^2+\|C_I\|^2}{\|C_I\|^2}
%\lambda_{12} & = \frac{M_{\textbf{\emph{C}}_R>0}(\textbf{\emph{C}}_R)+%M_{\textbf{\emph{C}}_I>0}(\textbf{\emph{C}}_I)}{M_{\textbf{\emph{C}}_R%>0}(\textbf{\emph{C}}_R)}\\
%\lambda_{22} & = \frac{M_{\textbf{\emph{C}}_R>0}(\textbf{\emph{C}}_R)+%M_{\textbf{\emph{C}}_I>0}(\textbf{\emph{C}}_I)}{M_{\textbf{\emph{C}}_I%>0}(\textbf{\emph{C}}_I)}\\
%u & = M_{\textbf{\emph{C}}_R>0}(\textbf{\emph{C}}_R)\\
%v & = M_{\textbf{\emph{C}}_I>0}(\textbf{\emph{C}}_I)
where $\textbf{\emph{C}}_F$ and $\textbf{\emph{B}}_F$ are respectively the fused results of the classification and regression branches. 
$\textbf{\emph{C}}_\textrm{RGB}$ and $\textbf{\emph{B}}_\textrm{RGB}$ are the outputs of the two  branches from the RGB images. 
Similarly, $C_\textrm{TIR}$ and $R_\textrm{TIR}$ are the outputs from the TIR images. 
$M$ denotes the 'mean' operation.
$\lambda_{12}$ and $\lambda_{22}$ are the learned weights, while $\lambda_{11}$ and $\lambda_{21}$ are fixed throughout the tracking process. 
$s$ is the scaling factor. 
The merits of our DFAT are confirmed by self-comparison and comparison with state-of-the-art trackers on the VOT-RGBT2019 dataset presented in the experimental section.

\subsection{Template Update}
As confirmed in \cite{UpdateNet}, updating the template is beneficial to tracking performance, especially in long-term scenarios. 
Traditionally, the target of each video sequence is annotated at the first frame and the appearance model is also learned from the initial frame. 
In principle, the target is often subject to appearance variation throughout the video sequence, which means a tracker with a constant template would encounter a model degradation. 
Consequently, with an attempt to make the template exploit temporal clues, in our approach, it is updated by linear interpolation after the object is located in the current frame. 
Specifically, for the embeddings of the three neck modules, we use a pre-defined learning rate to combine them with their corresponding historical template embeddings. 
% And the same operation is employed for the other two outputs from the leaving neck modules. 
% Note that the learning rate assigned to different necks is the same.
To prevent the template from over-fitting, we update the template every 10 frames.

\section{Experiments}\label{experiment}
In this section, we present the implementation details.
The performance of the fusion strategies developed at the three levels is measured on the VOT-RGBT2019 dataset. 
% Our implementation details are also discussed. 
In addition, we also verify the merits of the proposed DFAT on the published VOT-RGBT2020 and GTOT dataset. 

\subsection{Implementation Details}
\textbf{Experimental platform:} Our method DFAT is implemented with Pytorch 0.4.1 on a platform with Intel Core i9-9980XE CPU and NVIDIA GeForce RTX 2080Ti GPU. The speed of DFAT reaches 20 Frame Per Second (FPS).

\par \textbf{Parameters setup:} In our experiments, the hyperparameters $\lambda_{11}$ and $\lambda_{21}$ are set to 0.5 for fusing the results from the regression branches. 
To guarantee the scale continuity in the successive frames, we set the learning rate for updating the width and height of the bounding box to 0.32.
The learning rate for the linear interpolation update is chosen to be 0.1. 
The single-channel TIR images are triplicated to emulate RGB channels and thus our network can process TIR inputs without any modification. 
% All the parameters are fixed for all the experiments.

\begin{figure}
	\begin{center}
		\includegraphics[width=1\linewidth, height=0.6\linewidth]{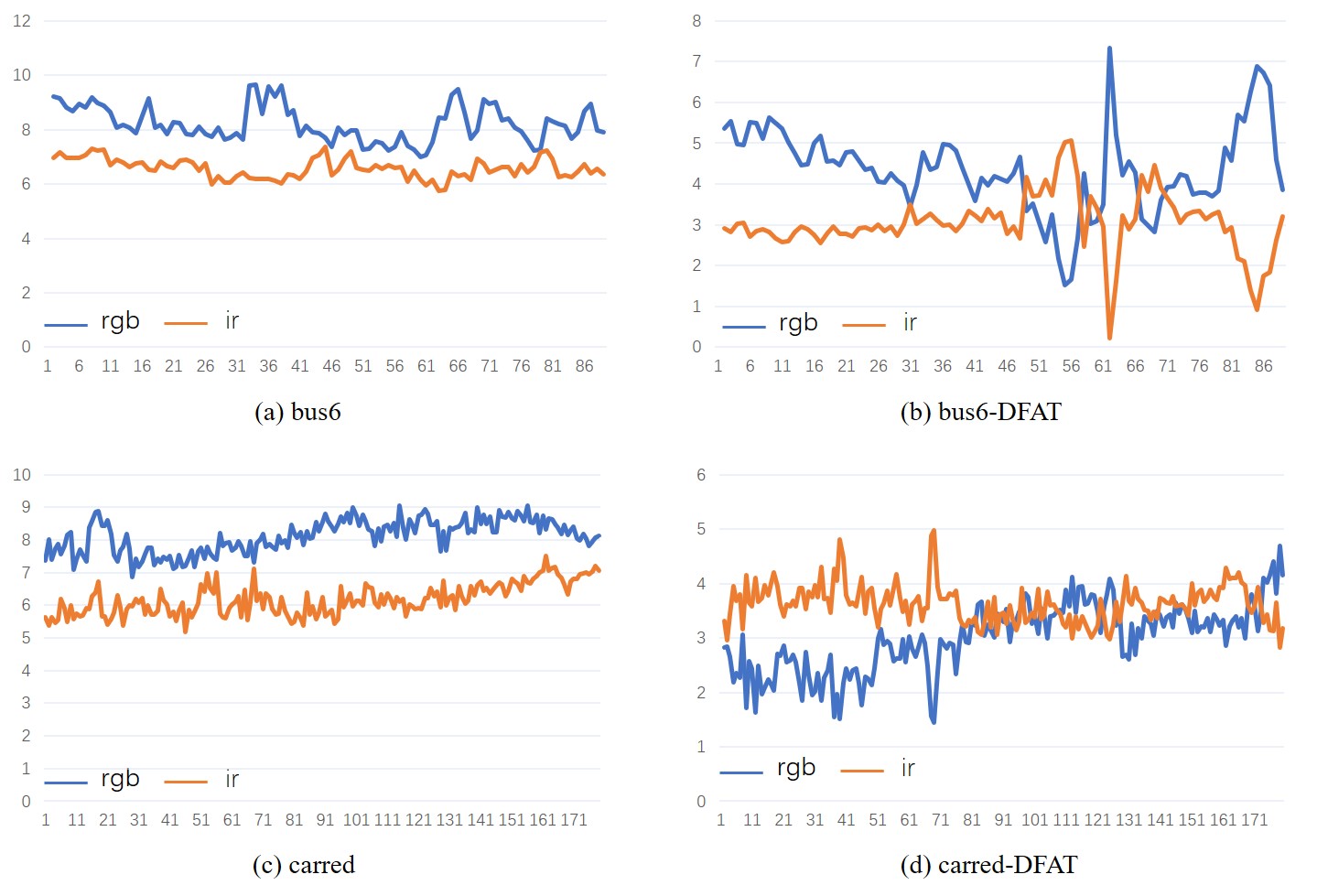} 
	\end{center}
	\caption{Visualization of the original classification scores and scores after debaising.}
	\label{bias-score}
\end{figure}

\subsection{Experimental Setup}
\textbf{Network:} Our backbone network (ResNet-50) \cite{Resnet50} is pretrained on ImageNet \cite{ImageNet}.
The RPN blocks and neck modules are trained by SGD optimizer on COCO \cite{COCO} and YouTube-VOS \cite{YouTube-VOS} with the batch size set to 96. 
A weight decay of 0.0001 and momentum of 0.9 are employed.
The learning rate is warmed up from 0.001 to 0.005 in the first 3 epochs and then it decays to 0.0025 in the remaining 17 epochs.
It is worth noting that the target mask in these two datasets is enabled, which contributes to the high accuracy of DFAT.

% \par \textbf{Datasets:} Our experiments are evaluated on the VOT-RGBT2019 \cite{VOT2019} dataset, which consists of 60 pairs of RGB and TIR video sequences. 
\textbf{Datasets:} Our experiments are evaluated on the VOT-RGBT2019 \cite{VOT2019}, VOT-RGBT2020 \cite{VOT2020} and GTOT datasets.

On VOT-RGBT2019 and VOT-RGBT2020, which are consisted of 60 video pairs, we use three evaluation measures named accuracy (A), robustness (R), and Expected Average Overlap (EAO) to estimate the results. 
Accuracy is measured in terms of the overlap of the estimated bounding boxes with the ground truth, while robustness is inversely proportional to the failure rate. 
EAO is the key criterion for evaluating a tracker in the VOT protocol. 
Note that a re-start mechanism is employed when the overlap between the calculated bounding box and ground truth falls to zero. 
As the original protocol is prone to favour the trackers using a larger bounding box, a different evaluation protocol is adopted in the VOT2020 challenge \footnote{https://data.votchallenge.net/vot2020/vot-2020-protocol.pdf}. 
The reset mechanism is replaced by initialization points referred to as anchors. 
Trackers run forward if the start anchor is before the middle of the video and vice versa. 
All the anchors are set manually with an average gap of 50 frames.

GTOT includes 100 video sequences and 15700 frames in total.
Success and precision rate are the two evaluation metrics on this dataset.
Success rate is the major measurement which counts the percentage of successfully predicted frames, whose IoU between their corresponding given bounding boxes are above a certain threshold, 0.6 in the benchmark.
Likewise, a prediction is considered precise if the distance between its centre point and the groundtruth is below a threshold, 5 in the benchmark.
After the number of precise frames are obtained, the precision rate can be computed by dividing 15700.

\subsection{Results of Self-Comparison}
In this section, we validate our method by self-comparison. 
Three fusion mechanisms are discussed in detail. 
The results are displayed in TABLE \ref{results_pixel_table}, \ref{results_feature_table} and \ref{results_decision_table}, respectively.

\begin{table}[h]
	\renewcommand\arraystretch{1}
% 	\resizebox{\textwidth}{7mm}{
	\centering
	\caption{\label{results_pixel_table}The results of pixel-level fusion (Fig. \ref{pixel-level}) on VOT-RGBT2019 dataset with MDLatLRR employed as the fusion strategy. (The best result is highlighted in \bfseries bold)}
	\scalebox{0.8}{	
		\begin{tabular}{c|c|c|c|c|c|c|c|c|c|c}
			\toprule
			\multicolumn{2}{c}{\multirow{2}{*}{ Level}} & \multicolumn{2}{c}{\multirow{2}{*}{Input Images}} &
			\multicolumn{7}{c}{Results} \\ \cline{6-6} \cline{8-8} \cline{10-10}
			\multicolumn{1}{c}{} & \multicolumn{1}{c}{} & \multicolumn{1}{c}{} & \multicolumn{1}{c}{} & \multicolumn{1}{c}{} & \multicolumn{1}{c}{A ($\uparrow$)} & \multicolumn{1}{c}{} & \multicolumn{1}{c}{R ($\downarrow$)} & \multicolumn{1}{c}{} & \multicolumn{1}{c}{EAO ($\uparrow$)} & \multicolumn{1}{c}{} \\
			\toprule
			\multicolumn{2}{c}{--} & \multicolumn{2}{c}{TIR + TIR} &
			\multicolumn{1}{c}{} & \multicolumn{1}{c}{0.6248} & \multicolumn{1}{c}{} & \multicolumn{1}{c}{0.4531} & \multicolumn{1}{c}{} & \multicolumn{1}{c}{0.2826} & \multicolumn{1}{c}{} \\
			\midrule
			\multicolumn{2}{c}{--} & \multicolumn{2}{c}{RGB + RGB} &
			\multicolumn{1}{c}{} & \multicolumn{1}{c}{0.6228} & \multicolumn{1}{c}{} & \multicolumn{1}{c}{0.4481} & \multicolumn{1}{c}{} & \multicolumn{1}{c}{0.3189} & \multicolumn{1}{c}{} \\
			\midrule
			\multicolumn{2}{c}{--} & \multicolumn{2}{c}{RGB + TIR} &
			\multicolumn{1}{c}{} & \multicolumn{1}{c}{\bfseries0.6482} & \multicolumn{1}{c}{} & \multicolumn{1}{c}{\bfseries0.3784} & \multicolumn{1}{c}{} & \multicolumn{1}{c}{0.3433} & \multicolumn{1}{c}{} \\
			\midrule
			\multicolumn{2}{c}{2} & \multicolumn{2}{c}{Fused + Fused} & 
			\multicolumn{1}{c}{} & \multicolumn{1}{c}{0.6180} & \multicolumn{1}{c}{} & \multicolumn{1}{c}{0.4332} & \multicolumn{1}{c}{} & \multicolumn{1}{c}{0.3120} & \multicolumn{1}{c}{} \\
			\midrule
			\multicolumn{2}{c}{3} & \multicolumn{2}{c}{Fused + Fused} & 
			\multicolumn{1}{c}{} & \multicolumn{1}{c}{0.6199} & \multicolumn{1}{c}{} & \multicolumn{1}{c}{0.4132} & \multicolumn{1}{c}{} & \multicolumn{1}{c}{0.3216} & \multicolumn{1}{c}{} \\
			\midrule
			\multicolumn{2}{c}{2} & \multicolumn{2}{c}{Fused + TIR} & 
			\multicolumn{1}{c}{} & \multicolumn{1}{c}{0.6138} & \multicolumn{1}{c}{} & \multicolumn{1}{c}{0.4182} & \multicolumn{1}{c}{} & \multicolumn{1}{c}{0.3286} & \multicolumn{1}{c}{} \\
			\midrule
			\multicolumn{2}{c}{3} & \multicolumn{2}{c}{Fused + TIR} & 
			\multicolumn{1}{c}{} & \multicolumn{1}{c}{0.6262} & \multicolumn{1}{c}{} & \multicolumn{1}{c}{0.3933} & \multicolumn{1}{c}{} & \multicolumn{1}{c}{\bfseries0.3481} & \multicolumn{1}{c}{} \\
			\bottomrule
		\end{tabular}
	}
% 	}
	
\end{table}

\par
\begin{figure}
	\begin{center}
		\includegraphics[width=1\linewidth]{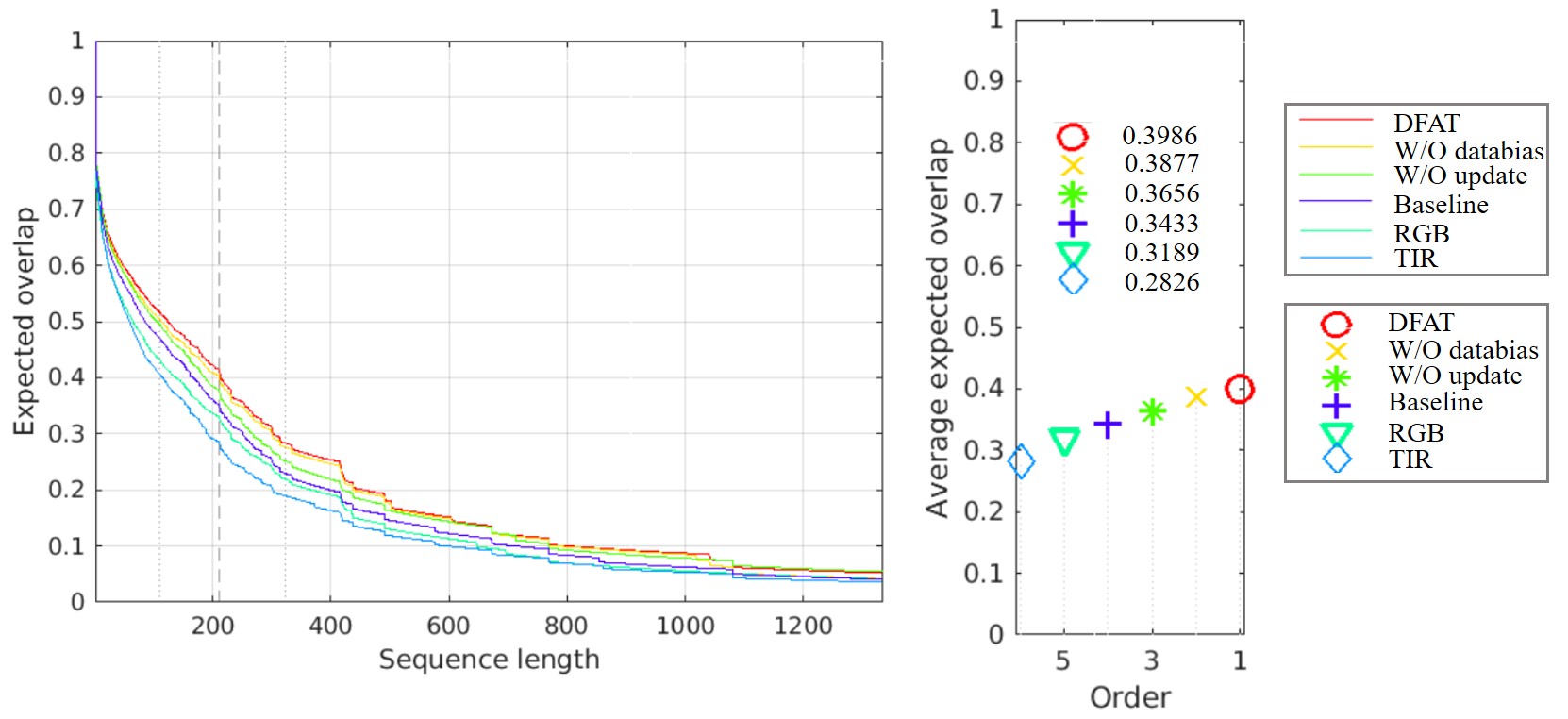} 
	\end{center}
	\caption{Expected overlap as a function of sequence length and the average expected overlap (EAO) of all variants.}
	\label{ablation study1}
\end{figure}

 \textbf{Pixel-level fusion: }For pixel-level fusion, as described in Fig. \ref{pixel-level}, we choose the fused images from the second and third decomposition level of MDLatLRR. 
The experimental results are reported in TABLE \ref{results_pixel_table}. 
Here 'fused' means the output image of MDLatLRR. 
As we can see, the results of images from level 3 are better than those of level 2. 
For these two levels, better performance is obtained by adding TIR images as the second input. 
The results of the baseline achieved with only TIR or RGB images reach 0.2826 and 0.3189 on the EAO evaluation metric, respectively. 
Thanks to the complementary information conveyed by the RGB and TIR modalities, the results of the pixel-level fusion by MDLatLRR show an improvement of 3.9\% and 0.27\%. 
The outputs from deeper levels of MDLatLRR are expected to capture the salient information from each modality, leading to better performance. 
Since fused images are visually close to the RGB modality, we also pair them with the original TIR image as the input of the tracking network and an improvement of 2.65\% on EAO is obtained.
However, as pixel-level fusion is not suitable for the end-to-end framework, 
the most corresponding results are lower than the result of using the original RGB and TIR images as inputs (Baseline), which is also shown in Fig. \ref{ablation study1}. 
\begin{table*}[ht]\footnotesize
	\renewcommand\arraystretch{1.2}
	\centering
	\caption{\label{results_feature_table}The results of fusion at feature-level on the VOT-RGBT2019 dataset. 'Selection' is the flag of whether using feature selection or not. Respectively, 'C' and 'S' mean channel attention and spatial attention. 'Vector' denotes the pooling method to obtain the attention vector. 'Scalar' represents the manner to acquire the fusion scalar from its corresponding attention vector. For example, without feature selection, the attention vector for the first line is calculated by average pooling spatially and then the fusion scalar is obtained by applying a maximum operator to this vector. (The best result is highlighted in \bfseries bold)}
	\scalebox{1}{
		\begin{tabular}{c|c|c|c|c|c|c|c|c|c|c}
			\toprule
			\multicolumn{1}{c}{\multirow{2}{*}{Selection}} &
			\multicolumn{1}{c}{\multirow{2}{*}{Type}} & \multicolumn{1}{c}{\multirow{2}{*}{Vector}} &
			\multicolumn{1}{c}{\multirow{2}{*}{Scalar}} &  \multicolumn{7}{c}{Results} \\ \cline{6-6} \cline{8-8} \cline{10-10}
			\multicolumn{1}{c}{} & \multicolumn{1}{c}{} &  \multicolumn{1}{c}{} & \multicolumn{1}{c}{} & \multicolumn{1}{c}{} & \multicolumn{1}{c}{A ($\uparrow$)} & \multicolumn{1}{c}{} & \multicolumn{1}{c}{R ($\downarrow$)} & \multicolumn{1}{c}{} & \multicolumn{1}{c}{EAO ($\uparrow$)} & \multicolumn{1}{c}{} \\
			\toprule
			\multicolumn{1}{c}{0} & \multicolumn{1}{c}{S} & \multicolumn{1}{c}{mean} & 
			\multicolumn{1}{c}{max} & \multicolumn{1}{c}{} & \multicolumn{1}{c}{\bfseries 0.6570} & \multicolumn{1}{c}{} & \multicolumn{1}{c}{0.3883} & \multicolumn{1}{c}{} & \multicolumn{1}{c}{0.3470} & \multicolumn{1}{c}{} \\
			\midrule
			\multicolumn{1}{c}{0} & \multicolumn{1}{c}{C} & \multicolumn{1}{c}{mean} & 
			\multicolumn{1}{c}{max} & \multicolumn{1}{c}{} & \multicolumn{1}{c}{0.6398} & \multicolumn{1}{c}{} & \multicolumn{1}{c}{0.3136} & \multicolumn{1}{c}{} & \multicolumn{1}{c}{0.3593} & \multicolumn{1}{c}{} \\
			\midrule
			\multicolumn{1}{c}{0} & \multicolumn{1}{c}{S} & \multicolumn{1}{c}{max} & 
			\multicolumn{1}{c}{mean} & \multicolumn{1}{c}{} & \multicolumn{1}{c}{0.6433} & \multicolumn{1}{c}{} & \multicolumn{1}{c}{0.3037} & \multicolumn{1}{c}{} & \multicolumn{1}{c}{0.3615} & \multicolumn{1}{c}{} \\
			\midrule
			\multicolumn{1}{c}{0} & \multicolumn{1}{c}{C} & \multicolumn{1}{c}{max} & 
			\multicolumn{1}{c}{mean} & \multicolumn{1}{c}{} & \multicolumn{1}{c}{0.6350} & \multicolumn{1}{c}{} & \multicolumn{1}{c}{0.3136} & \multicolumn{1}{c}{} & \multicolumn{1}{c}{0.3618} & \multicolumn{1}{c}{} \\
			\midrule
			\multicolumn{1}{c}{1} & \multicolumn{1}{c}{S} & \multicolumn{1}{c}{max} & 
			\multicolumn{1}{c}{mean} & \multicolumn{1}{c}{} & \multicolumn{1}{c}{0.6433} & \multicolumn{1}{c}{} & \multicolumn{1}{c}{0.3037} & \multicolumn{1}{c}{} & \multicolumn{1}{c}{0.3615} & \multicolumn{1}{c}{} \\
			\midrule
			\multicolumn{1}{c}{1} & \multicolumn{1}{c}{C} & \multicolumn{1}{c}{max} & 
			\multicolumn{1}{c}{mean} & \multicolumn{1}{c}{} & \multicolumn{1}{c}{0.6416} & \multicolumn{1}{c}{} & \multicolumn{1}{c}{\bfseries 0.2987} & \multicolumn{1}{c}{} & \multicolumn{1}{c}{\bfseries 0.3788} & \multicolumn{1}{c}{} \\
			\bottomrule
		\end{tabular}	
	}
\end{table*}

			%\midrule
			%\multicolumn{1}{c}{1} & \multicolumn{1}{c}{C} & \multicolumn{1}{c}{max} & 
			%\multicolumn{1}{c}{mean} & \multicolumn{1}{c}{} & \multicolumn{1}{c}{0.6465} & %\multicolumn{1}{c}{} & \multicolumn{1}{c}{0.3087} & \multicolumn{1}{c}{} & %\multicolumn{1}{c}{0.3667} & \multicolumn{1}{c}{} \\
			%\midrule
			%\multicolumn{1}{c}{1} & \multicolumn{1}{c}{S} & \multicolumn{1}{c}{max} & 
			%\multicolumn{1}{c}{mean} & \multicolumn{1}{c}{} & \multicolumn{1}{c}{0.6412} & %\multicolumn{1}{c}{} & \multicolumn{1}{c}{0.3037} & \multicolumn{1}{c}{} & %\multicolumn{1}{c}{0.3737} & \multicolumn{1}{c}{} \\

\par \textbf{Feature-level fusion: } It is expected that fusion at the feature level will deliver better performance than pixel-level fusion. 
This is confirmed in our study.
Generally, we compute the fusion weights from the attention vectors, which are previously obtained by the attention mechanism.
Specifically, for fusion with feature selection, we first employ global max pooling to omit the channels with lower discriminative power and keep the sequential order immutably ('Selection'). 
Considering the feature noise and redundancy, we set the selection rate to 20\% so that 80\% of the channels are retained. 
After the type of attention to be used is selected ('Type'), we obtain the corresponding attention vector with average or max-pooling ('Vector'). 
And the representative scalar is supposed to be calculated from its corresponding attention vector ('Scalar'), which is used for feature-level fusion after normalization.
%At this step, 'mean' represents GAP for channel attention and spatial attention. 
%We quantify the importance of the feature ('Choice') by either the mean or the max value of the attention map. 
%The filter operation denotes fusion without feature selection, which is referred to as '-'. 
% All the experiments in TABLE \ref{results_feature_table} are executed with an update. 
It can be seen that by executing feature-level fusion after removing less discriminative channels, we get the best result in robustness (0.2987) as well as EAO (0.3788) at this fusion level.

\begin{table*}[ht]
	\renewcommand\arraystretch{1.2}
	\footnotesize
	\centering
	\caption{\label{results_decision_table}A comparison of the decision-level fusion before and after normalization on the VOT-RGBT2019 dataset. Here 'DF' means the decision-level fusion strategy proposed in this paper while 'KLD' represents KL Divergence. 'After' and 'Before' separately denote fusion after and before normalization. (The best result is highlighted in \bfseries bold) }
	\scalebox{1}{	
		\begin{tabular}{c|c|c|c|c|c|c|c|c|c|c|c|c}
			\toprule
			\multicolumn{1}{c}{\multirow{2}{*}{Operating bias}} &
			\multicolumn{1}{c}{\multirow{2}{*}{Location}} & \multicolumn{5}{c}{Fusion Strategy} &
			\multicolumn{6}{c}{Results} \\ \cline{4-4} \cline{6-6} \cline{8-8} \cline{10-10} \cline{12-12}
			\multicolumn{1}{c}{} & \multicolumn{1}{c}{} &  \multicolumn{1}{c}{} & \multicolumn{1}{c}{Cls} & \multicolumn{1}{c}{} & \multicolumn{1}{c}{Reg} & \multicolumn{1}{c}{} & \multicolumn{1}{c}{A ($\uparrow$)} & \multicolumn{1}{c}{} & \multicolumn{1}{c}{R ($\downarrow$)} & \multicolumn{1}{c}{} & \multicolumn{1}{c}{EAO ($\uparrow$)} & \multicolumn{1}{c}{} \\
			\toprule
			\multicolumn{1}{c}{F} & \multicolumn{1}{c}{After} & \multicolumn{1}{c}{} & 
			\multicolumn{1}{c}{KLD} & \multicolumn{1}{c}{} & \multicolumn{1}{c}{KLD} & \multicolumn{1}{c}{} & \multicolumn{1}{c}{0.6523} & \multicolumn{1}{c}{} & \multicolumn{1}{c}{0.3385} & \multicolumn{1}{c}{} & \multicolumn{1}{c}{0.3680} & \multicolumn{1}{c}{}\\
			\midrule
			\multicolumn{1}{c}{F} & \multicolumn{1}{c}{After} & \multicolumn{1}{c}{} & 
			\multicolumn{1}{c}{KLD} & \multicolumn{1}{c}{} & \multicolumn{1}{c}{0.5} & \multicolumn{1}{c}{} & \multicolumn{1}{c}{\bfseries0.6667} & \multicolumn{1}{c}{} & \multicolumn{1}{c}{0.3435} & \multicolumn{1}{c}{} & \multicolumn{1}{c}{0.3656} & \multicolumn{1}{c}{}\\
			\midrule
			\multicolumn{1}{c}{F} & \multicolumn{1}{c}{After} & \multicolumn{1}{c}{} & 
			\multicolumn{1}{c}{0.5} & \multicolumn{1}{c}{} & \multicolumn{1}{c}{0.5} & \multicolumn{1}{c}{} & \multicolumn{1}{c}{0.6648} & \multicolumn{1}{c}{} & \multicolumn{1}{c}{0.2788} & \multicolumn{1}{c}{} & \multicolumn{1}{c}{0.3863} & \multicolumn{1}{c}{}\\		
			\midrule
			\multicolumn{1}{c}{F} & \multicolumn{1}{c}{Before} & \multicolumn{1}{c}{} & 
			\multicolumn{1}{c}{0.5} & \multicolumn{1}{c}{} & \multicolumn{1}{c}{0.5} & \multicolumn{1}{c}{} & \multicolumn{1}{c}{0.6619} & \multicolumn{1}{c}{} & \multicolumn{1}{c}{0.2788} & \multicolumn{1}{c}{} & \multicolumn{1}{c}{0.3877} & \multicolumn{1}{c}{}\\
			\midrule
			\multicolumn{1}{c}{T} & \multicolumn{1}{c}{Before} & \multicolumn{1}{c}{} & 
			\multicolumn{1}{c}{DF} & \multicolumn{1}{c}{} & \multicolumn{1}{c}{DF} & \multicolumn{1}{c}{} & \multicolumn{1}{c}{0.5978} & \multicolumn{1}{c}{} & \multicolumn{1}{c}{0.2639} & \multicolumn{1}{c}{} & \multicolumn{1}{c}{0.3663} & \multicolumn{1}{c}{}\\
			\midrule
			\multicolumn{1}{c}{T} & \multicolumn{1}{c}{Before} & \multicolumn{1}{c}{} & 
			\multicolumn{1}{c}{DF} & \multicolumn{1}{c}{} & \multicolumn{1}{c}{0.5} & \multicolumn{1}{c}{} & \multicolumn{1}{c}{0.6652} & \multicolumn{1}{c}{} & \multicolumn{1}{c}{\bfseries0.2539} & \multicolumn{1}{c}{} & \multicolumn{1}{c}{\bfseries0.3986} & \multicolumn{1}{c}{}\\
			\bottomrule
		\end{tabular}
	}
\end{table*}

\par \textbf{Decision-level fusion: }For the decision-level fusion, we consider two variants, depending on the positioning of the fusion step, \ie before normalization ('Before') and after normalization ('After'). 
% The data bias between RGB and TIR modalities, as mentioned above, is alleviated by normalization.
Since the classification scores are restrainted between 0 and 1 after normalization, the data bias between RGB and TIR modalities vanishes.
% That means fusion after normalization has a tendency inherently. 
We compare the fusion results at these two variants to confirm our hypothesis. 
% Besides, the experiments at this level are all with the update. 
As mentioned before, another tracker uses decision-level fusion to obtain a reliable response map with KL Divergence \cite{kld}. 
This criterion is well suited for a probabilistic model matching, exemplified by the case of fusion after normalization. 
On VOT-RGBT2019, the given ground truth comes from the TIR modality. 
As the paired RGB and TIR images are not accurately aligned, the misalignment increases the difficulty of handling the problem of fusion in the regression branch. 
So we just consider exploiting both KL Divergence and our method on the classification branch. 
As can be seen, the variant with the KL Divergence used after normalization does not achieve a promising result, which is even worse than adding the two sources of information together with the weights set to 0.5. 
This is caused by the labels used in network training not being of a Gaussian-shape matrix, whereas \cite{kld} is a CF-based tracker, with the target response satisfying this requirement. 
Thus, KL Divergence performs unsatisfactorily in our framework.

On the other hand, the fusion before normalization is more effective which also indicates the necessity of debiasing. 
The result with weights manually set to 0.5 is 0.3877 on EAO, achieving a gain of 0.14\% compared to the highest result after normalization. 
By implanting our method on the classification branch, the result of our tracker, which ranks the best in the self-comparison, reaches 0.3986. 
We obtain a performance improvement of 1.09\% on EAO and 2.49\% in failure rate.

\par \textbf{Illustrating the details of the decision-level fusion: }In order to gain a clear understanding of our decision-level fusion, we first clarify the three numerical operations, \ie \emph{adaptive weighting}, \emph{normalization} and \emph{scaling}. 
As illustrated in the second part of Section III, adaptive weighting method is applied to reduce the bias, which is visualized in Fig. \ref{bias-score}\textcolor{red}{(a)} and \ref{bias-score}\textcolor{red}{(c)}, as well as make better use of the complementary information from different modalities. 
We use softmax as the normalization method as most Siamese-based trackers do.
Actually, the fused result does not represent the final measure of 'similarity', which is of great importance in the Siamese framework \cite{SiamFC}. 
From the fused result, we do not know whether the candidate anchor is considered positive or not.
So the normalization operation is introduced to convert the raw results to probabilities, which is also consistent with the subsequent operations in the framework.
The scaling factor is instrumental in achieving better results in the classification branch, which is discussed in the fifth part of Section IV.

\par The experiments demonstrate that fusion at the decision level is the best mechanism in an end-to-end framework trained without TIR images. 
Ignoring the method proposed in this paper, a simple fusion at the decision level achieves 0.3877, which is merely worse than DFAT. 
Besides, fusion before normalization is more effective than fusion after normalization. 
Moreover, from Fig. \ref{bias-score}\textcolor{red}{(b)} and \ref{bias-score}\textcolor{red}{(d)}, it can be seen that the bias is significantly reduced after equipped with our debiasing module. 

\subsection{Comparing to State-of-the-art Trackers}
In this section, we verify the merits of our method by comparing it to the state-of-the-art trackers. 
Here the comparative trackers include JMMAC \cite{JMMAC}, SiamDW\_T \cite{SiamDW}, CISRDCF, FSRPN, GESBTT, MPAT, mfDimp \cite{mfDimp}, MANet \cite{MANet}, FANet \cite{FANet} and TFNet \cite{TFNet}. 
All the trackers can be found in \cite{VOT2019} except the last two trackers. 
%As the robustness displayed in \cite{VOT2019} is different from the value obtained from the vot-toolkit, we only compare the EAO and accuracy of the trackers in this paper. 
TABLE \ref{compare_sota_table} provides the results of trackers on VOT-RGBT2019 dataset. 
Our DFAT ranks marginally the best in accuracy and the second in terms of EAO. 
RPN is utilized for precise estimation of the bounding box, which proves that FSRPN and DFAT are both among the top three trackers in accuracy. 
As can be seen, a compact mask is a more exact requirement than the bounding box. 
So training the network based on the masks also contributes to the high accuracy of DFAT. 
DFAT only falls behind the first place of the published VOT-RGBT2019 dataset on EAO evaluation and exceeds the runner-up of about 0.61\%.
As the robustness displayed in \cite{VOT2019} is different from the value obtained from the vot-toolkit, for completeness, the exhibited scores for all contrastive trackers are received by one minus the published scores in \cite{VOT2019} and \cite{TFNet}.

\begin{figure}
	\begin{center}
		\includegraphics[width=1\linewidth]{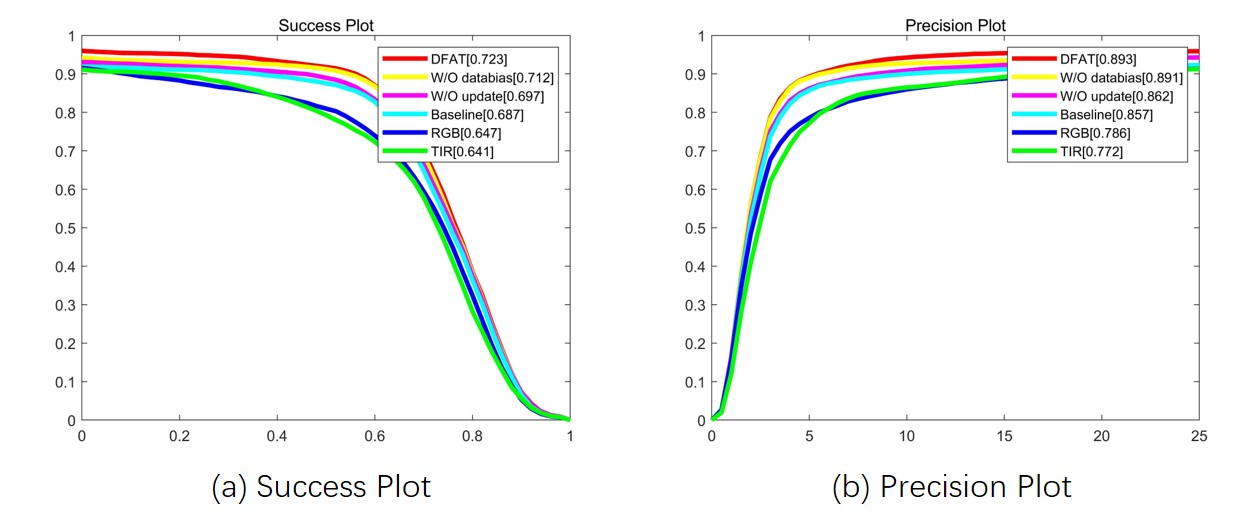} 
	\end{center}
	\caption{The success and precision rates of all variants on GTOT.}
	\label{ablation study2}
\end{figure}

\begin{figure*}
	\begin{center}
		\includegraphics[width=0.95\linewidth]{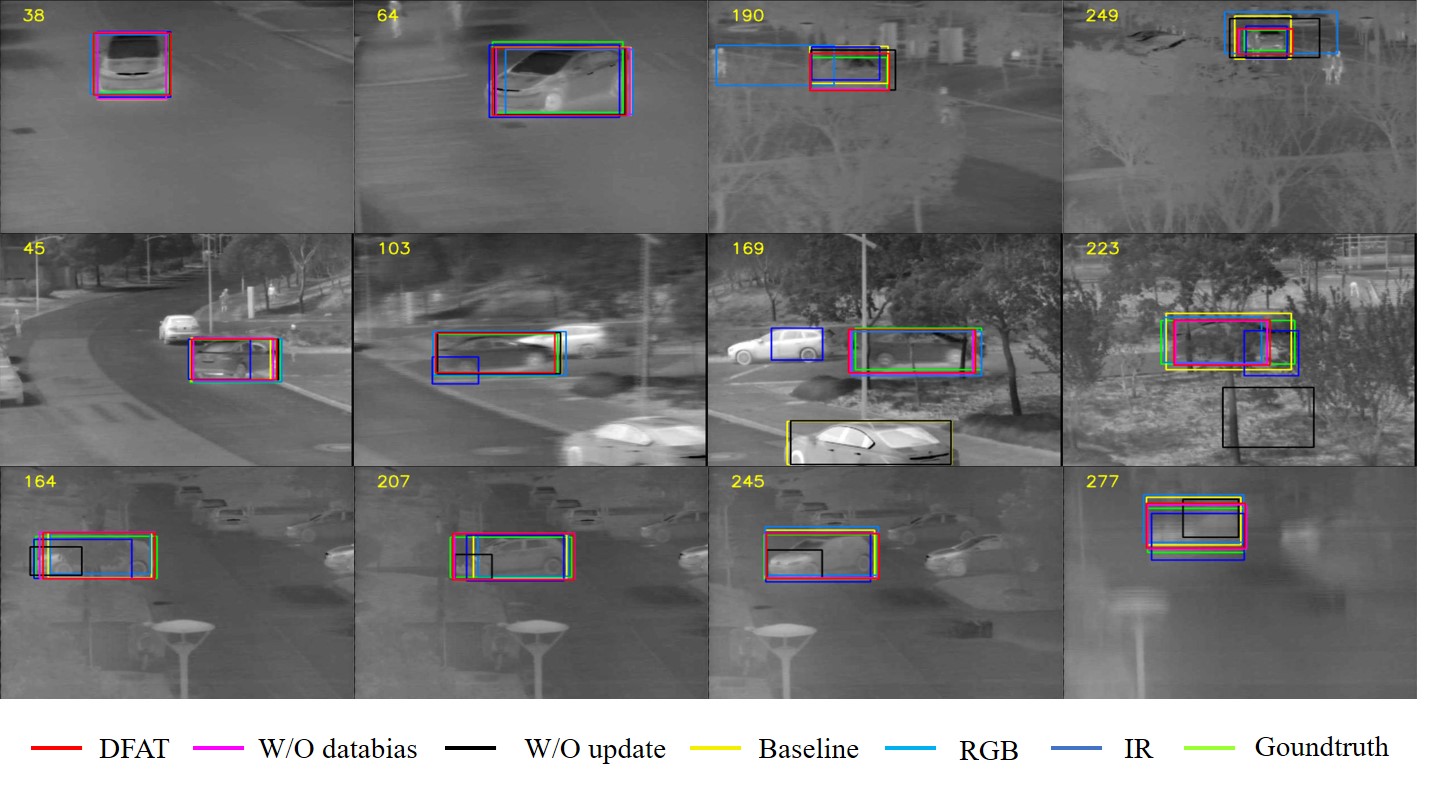} 
	\end{center}
	\caption{A subjective comparison of the respective trackers on the VOT-RGBT2020 dataset. From top to bottom, we show the results on three video sequences, \ie  \emph{car41}, \emph{caraftertree} and \emph{carnotfar}.}
	\label{results_image_tir}
\end{figure*}

\subsection{Self-Analysis}

\textbf{Ablation study:} In this subsection, we perform an analysis of the impact of each component of our DFAT tracker, \ie template update and adaptive fusion at the decision level. 
The visualization is presented in Fig. \ref{ablation study1} and Fig. \ref{ablation study2}, in which 'W/O bias' is the tracker without bias correction and 'W/O update' represents without the update.
'Baseline' means the extension of our baseline tracker SiamRPN++ to RGBT tracking while RGB and TIR mean only RGB and TIR images available.
Respectively, 'RGB' and 'TIR' means only the visible and thermal infrared data is used as input.

According to the ablation study on the VOT-RGBT2019 dataset, the trackers with single modality information perform undesirable, with the result on RGB being 0.3189 in terms of EAO.
Since our network is trained on RGB datasets, using it for tracking on the TIR input achieves an even worse result, the EAO of which is 0.2826. 
By exploiting the complementary information between the two modalities, the baseline RGBT tracker improves from 0.3189 to 0.3433. 
In the baseline tracker, a fixed template is employed while the appearance of the target will change continuously throughout the whole video sequence. 
Introducing a simple yet effective update strategy, which is realized by linear interpolation, we generate a more robust template and the results gain a 4.44\% enhancement compared to our baseline. 
By dealing with the preconception, significantly caused by the method of training, the 'W/O update' tracker outperforms our baseline by 2.23\%. 
An intuitive payoff is displayed in Fig. \ref{bias-score}. 
As can be seen, the scores of the RGB modality are not consistently higher than that of the TIR modality. 
The specific value will be discussed in the next subsection. 
Finally, by taking advantage of these two components jointly, the best result is achieved, with gains of 1.13\% and 3.30\% compared to 'W/O bias' and 'W/O update' respectively. 
The experimental results demonstrate the merits of the template update and reducing data bias.

Similarly, on GTOT, the success rate of RGB only tracker is also higher than the TIR only variant.
With multi-modal clues integrated, the baseline method obtains a success rate of 0.687.
Furthermore, the update and debiasing operators bring an increment of 2.5\% and 1\% respectively.
Jointly, our DFAT reaches 0.723 on the measurement of success rate and gains 3.6\% in total compared with baseline method.

%\textbf{Analysis of the scaling factor:} The adaptive fusion strategy proposed in this paper is carried out before normalization (softmax in this paper).

\begin{figure}
	\begin{center}
		\includegraphics[width=0.9\linewidth]{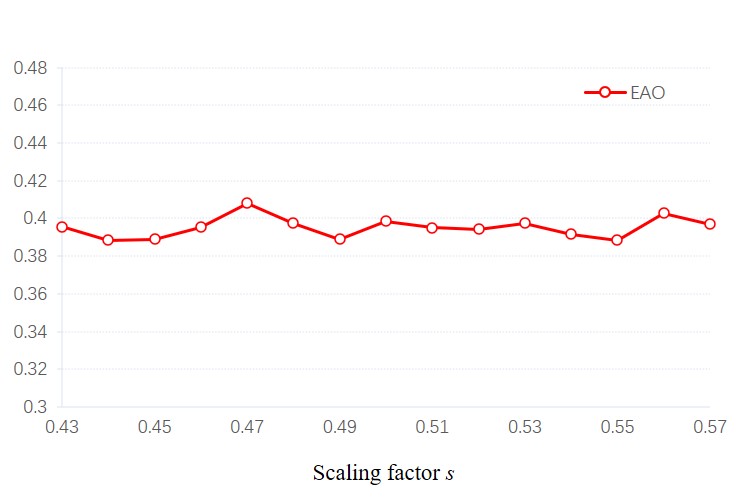} 
	\end{center}
	\caption{Analysis of the scaling factor on the VOT-RGBT2019 dataset.}
	\label{analysis of sclaing factor}
\end{figure}

\textbf{Analysis of the scaling factor:} Generally, as mentioned above, the results from the classification branch is transferred to $[0,1]$ after normalization (softmax in this paper).
Under this situation, an incorrect index would be received from the classification branch if there exists more than one large value (1) in the normalized scores, caused by rounding errors. 
This would result in a rough estimate of the bounding box.
To address this problem, we introduce a scaling factor to modulate the scores before normalization and amplify the disparity of the normalized scores. 
This is the reason why the scores handled by our method are lower than the original scores, as shown in Fig. \ref{bias-score}.
To illuminate further the impact of the scaling factor, a sensitivity analysis is performed on the VOT-RGBT2019 dataset. 
Fig. \ref{analysis of sclaing factor} reports the experimental results. 
With the fusion weights at the regression branch set to 0.5, we find that the worst result (0.3884) is obtained with a scaling factor of 0.44, while the tracker with a scaling factor of 0.47 ranks the best (0.4080). 
From the results shown, we can conclude that our DFAT is quite insensitive to the value of the scaling factor.

\begin{table*}[ht]
	\renewcommand\arraystretch{1.5}
	\centering
	\caption{\label{compare_sota_table}Tracking results on VOT-RGBT2019 dataset. (The best result is highlighted in \bfseries bold)}
	\scalebox{1}{
	\resizebox{\textwidth}{!}{
			\begin{tabular}{cccccccccccc}
			\toprule
			Tracker & FANet\cite{FANet} & TFNet\cite{TFNet} & GESBTT\cite{VOT2019} & CISRDCF\cite{VOT2019} & MPAT\cite{VOT2019} & MANet\cite{MANet} & FSRPN\cite{VOT2019} & mfDimp\cite{mfDimp} & SiamDW\_T\cite{SiamDW} & JMMAC\cite{JMMAC} & DFAT \\
			\toprule
			Published & TIV2021 & TCSVT2021 & - & - & - & ICCVW2019 & - & ICCVW2019 & CVPR2019 & TIP2021 & - \\
			\midrule
			A & 0.4724 & 0.4617 & 0.6163 & 0.5215 & 0.5723 & 0.5823 & 0.6362 & 0.6019 & 0.6158 & 0.6649 & \bfseries0.6652 \\
			\midrule
			R & 0.4922 & 0.4064 & 0.3650 & 0.3096 & 0.2758 & 0.2990 & 0.2931 & 0.1964 & 0.2161 & \bfseries0.1789 & 0.2539 \\
			\midrule
			EAO & 0.2465 & 0.2878 & 0.2896 & 0.2923 & 0.3180 & 0.3463 & 0.3553 & 0.3879 & 0.3925 & \bfseries0.4826 & 0.3986\\
			\bottomrule
		\end{tabular}
	}
	}
\end{table*}

\begin{table}[ht]\footnotesize
	\renewcommand\arraystretch{1.2}
	\centering
	\caption{\label{ablation_study_vot2020_table}Tracking results on VOT-RGBT2020 dataset. (The best result is highlighted in \bfseries bold)}
	\scalebox{1}{	
		\begin{tabular}{ccccc}
			\toprule
			Variation & Baseline & W/O update &W/O bias & DFAT \\
			\toprule
			A & 0.6751 & 0.6741 & 0.6754 & \bfseries0.6796\\
			\midrule
			R & 0.7423 & 0.7415 & 0.7728 & \bfseries0.7869\\
			\midrule
			EAO & 0.3902 & 0.3937 & 0.4073 & \bfseries0.4178\\
			\bottomrule
		\end{tabular}
	}
\end{table}

\subsection{Evaluation on the VOT-RGBT2020 Dataset}
To validate the merits of our method, we also examine DFAT on the VOT-RGBT2020 dataset \cite{VOT2020} with vot-toolkit 0.2.0.
Each proposed component is also reconfirmed to be effective. 
TABLE \ref{ablation_study_vot2020_table} provides an intuitive comparison for each component. 
Interestingly, the variant 'W/O bias' submitted to the VOT-RGBT2020 challenge won the championship.
Correcting the data bias, we gain an improvement of 1.05\% which raises the best result to 0.4178 and sets a new state-of-the-art in RGBT tracking. 
We also conducted qualitative experiments to assess the impact of each component and visualize the results in Fig. \ref{results_image_tir}. 
We show the results on TIR images for clarity. 
As is evident, our DFAT achieves the top performance. 
The 'W/O bias' variant is only marginally worse than DFAT and ranks second, which demonstrates the necessity of the template update. 
As for the experiments in TABLE \ref{ablation_study_vot2020_table}, the scaling factor is still set to 0.5. 
A more suitable scaling factor can be found to obtain better performance. 
For example, setting the scaling factor to 0.47, which is found to be the best on the VOT-RGBT2019 dataset, gains a slight increase of 0.04\%.

\section{Conclusion}\label{conclusion}
%In this paper, we proposed a novel fusion strategy at the decision-level for accurate multimodal (RGB-TIR) object tracking (DFAT) in video. 
In this paper, we first explore different fusion strategies at three levels, \ie pixel-level, feature-level and decision-level, and the experimental results show that fusion at the decision level performs the best with only visible data employed for training.
Therefore, we proposed a novel fusion strategy at the decision level for accurate multi-modal (RGB and TIR) object tracking (DFAT) in a video sequence.
The tracker is built using the SiamRPN++ architecture trained with RGB data only, but deployed for embedding both, the RGB and TIR modalities. 
As our network is trained using only RGB datasets, TIR images are copied three times to emulate the network input format of the RGB channels. 
The tracking is facilitated by a template updating mechanism based on a linear target interpolation every 10 frames. 

Based on a deep analysis of the outputs of the RPN blocks of SiamRPN++, we design a novel fusion strategy to merge the complementary information conveyed by the RGB and TIR modalities. It involves a dynamic de-biasing of the RGB and TIR contributions to the classification branch of the network. The strategy also performs a classification branch scaling to balance the relative contributions of the classification and regression results. A variant of the proposed decision-level fusion was the winning entry of the VOT-RGBT(2020) Challenge.
% Fusion at pixel-level and feature-level are also expounded for a comprehensive comparison. 
% At pixel-level, MDLatLRR is used to merge the RGB and TIR images by maintaining the details from both images. 
% Attention mechanism, as well as feature denoise and deredundancy operations, are applied at feature level. 
The experimental results on the VOT-RGBT2019, VOT-RGBT2020 and GTOT datasets confirm the effectiveness and robustness of our method. 
Moreover, the outcome of the VOT-RGBT2020 challenge also demonstrates that our DFAT represents a new state-of-the-art in RGBT tracking. 
% For future work, it's of great importance to enhance the representation capability for TIR images of the network. And in this paper, we come up with the fusion strategy without taking the regression branch into account. So we also believe that taking full use of the outputs of RPN blocks will enhance the accuracy further.

%% References


% or


%\appendices

%\section{}

\section*{Acknowledgement}
This work was supported in part by the National Key Research and Development Program of China under Grant 2017YFC1601800, in part by the National Natural Science Foundation of China (Grant NO.62020106012, U1836218, 61672265), the 111 Project of Ministry of Education of China (Grant No.B12018), and the UK EPSRC (EP/N007743/1, MURI/EPSRC/DSTL, EP/R018456/1).
% This work was supported in part by the EPSRC Programme Grant (FACER2VM) EP/N007743/1, EPSRC/dstl/MURI project EP/R018456/1, the National Natural Science Foundation of China (61373055, 61672265, 61876072, 61602390) and the NVIDIA GPU Grant Program.

% \ifCLASSOPTIONcaptionsoff
%   \newpage
% \fi



\bibliographystyle{IEEEtran}
\bibliography{sample-bib}
%\begin{thebibliography}{1}

\newpage

\end{document}